\documentclass[letterpaper, 10 pt, conference]{ieeeconf} 
\IEEEoverridecommandlockouts                             
\usepackage[utf8]{inputenc}
\usepackage[T1]{fontenc}
\usepackage{epsfig}
\usepackage{amsmath}
\usepackage{amssymb}
\usepackage{mathtools}
\let\labelindent\relax
\usepackage{enumitem}
\usepackage{flushend}
\usepackage{cite}
\usepackage{hyperref}
\usepackage{dblfloatfix}

\newcommand{\shrinka}{\def\baselinestretch{0.993}\large\normalsize}

\title{\LARGE \bf Sim-to-Real for Robotic Tactile Sensing via \\Physics-Based Simulation and Learned Latent Projections}
\author{Yashraj Narang*$^{1}$, Balakumar Sundaralingam*$^{1}$, Miles Macklin$^{1}$, Arsalan Mousavian$^{1}$, Dieter Fox$^{1, 2}$
\thanks{*These authors contributed equally.} \thanks{$^{1}$NVIDIA Corporation, Seattle, USA. $^{2}$Paul G. Allen School of Computer Science \& Engineering, University of Washington, Seattle, USA.}%
}

\begin{document}

\setcounter{figure}{1}
\makeatletter
\let\@oldmaketitle\@maketitle
\renewcommand{\@maketitle}{\@oldmaketitle
\begin{center}
    \centering     
    \includegraphics[width=0.9\textwidth]{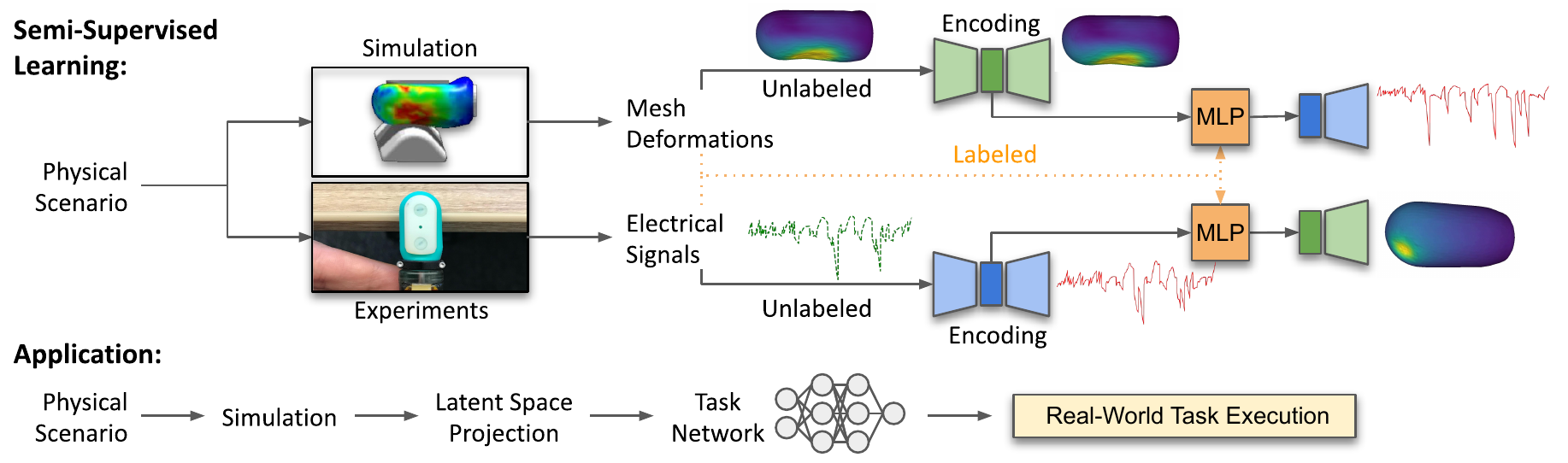}
\end{center}
  \footnotesize{\textbf{Fig.~\thefigure:\label{fig:modality_map}}~Overview. We develop an efficient 3D FEM model of a SynTouch BioTac sensor to simulate contact interactions, and we conduct similar real-world experiments. In a learning phase, we train autoencoders to reconstruct unlabeled FEM deformations and real-world electrical signals. With a small amount of labeled data, we subsequently train MLPs to project between the FEM and electrical latent spaces. At test time, we use these learned latent projections to perform cross-modal transfer between FEM and electrical data for unseen contact interactions. During downstream application, we 1) accurately synthesize BioTac electrical signals, and 2) estimate the shape and location of contact patches, facilitating real-world task execution.}\vspace{-18pt}
  \medskip}
\makeatother
\shrinka
\maketitle
\shrinka
\maketitle
\thispagestyle{empty}
\pagestyle{empty}

\begin{abstract}

Tactile sensing is critical for robotic grasping and manipulation of objects under visual occlusion. However, in contrast to simulations of robot arms and cameras, current simulations of tactile sensors have limited accuracy, speed, and utility. In this work, we develop an efficient 3D finite element method (FEM) model of the SynTouch BioTac sensor using an open-access, GPU-based robotics simulator. Our simulations closely reproduce results from an experimentally-validated model in an industry-standard, CPU-based simulator, but at 75x the speed. We then learn latent representations for simulated BioTac deformations and real-world electrical output through self-supervision, as well as projections between the latent spaces using a small supervised dataset. Using these learned latent projections, we accurately synthesize real-world BioTac electrical output and estimate contact patches, both for unseen contact interactions. This work contributes an efficient, freely-accessible FEM model of the BioTac and comprises one of the first efforts to combine self-supervision, cross-modal transfer, and sim-to-real transfer for tactile sensors.
\end{abstract}

\section{INTRODUCTION}

Tactile sensing is critical for grasping and manipulation under visual occlusion, as well as for handling delicate objects \cite{Billard2019Science}. For example, humans leverage tactile sensing when retrieving keys, striking a match, holding a wine glass, and grasping fresh fruit without damage. In robotics, researchers are actively developing a wide variety of tactile sensors (e.g., \cite{SynTouch, Wettels2008AdvRob, Yuan2017Sensors, Ward2018SoRo, Lambeta2020ICRA, Alspach2019RoboSoft, Padmanabha2020ICRA, Sferrazza2019Sensors, Yamaguchi2016Humanoids, Huang2019ICRA, McInroe2018IROS, Piacenza2020TMech}). These sensors have been used for tasks such as slip detection\cite{Su2015Humanoids, veiga2018grip, James2020RAL}, object classification\cite{hoelscher2015evaluation, Yuan2018ICRA}, parameter estimation\cite{Sundaralingam2020ArXiv}, 
force estimation\cite{Ma2018ICRA, Sundaralingam2019ICRA, Narang2020RSS, Narang2021arXiv}, contour following\cite{Lepora2019RAL}, and reacting to humans\cite{Huang2020ICRA}.

For other aspects of robotics, such as robot kinematics, dynamics, and cameras, accurate and efficient simulators have advanced the state-of-the-art in task performance. For example, simulators have enabled accurate testing of algorithms for perception, localization, planning, and control\cite{Afzal2020arXiv}; generation of synthetic datasets for learning such algorithms\cite{Tremblay2018ICRA, Matl2020ICRA, Matl2020CoRL}; efficient training of control policies via reinforcement learning (RL)\cite{Yu-RSS-17, Chebotar2019ICRA, OpenAI2019ArXiv}; and execution of online algorithms, with the simulator as a model~\cite{lowrey_simrl}. These capabilities have in turn reduced the need for costly, time-consuming, dangerous, or intractable experiments.

However, simulators for tactile sensing are still nascent. For the SynTouch BioTac sensor\cite{SynTouch, Wettels2008AdvRob}, as well as vision-based tactile sensors, most simulation studies have focused on the \textit{inverse} problem of interpreting sensor output in terms of physical quantities (e.g., \cite{Ma2018ICRA, Sferrazza2019Access, Narang2020RSS, Narang2021arXiv}). Far fewer efforts have addressed the \textit{forward} problem of synthesizing sensor output, and perhaps none have accurately generalized to diverse contact scenarios. Forward simulation is invaluable for simulation-based training, which coupled with domain adaptation, can enable effective policy generation.

The dearth of tactile simulation capabilities is a result of its inherent challenges. Accurate tactile sensor simulators must model numerous contacts, complex geometries, and elastic deformation, which can be computationally prohibitive\cite{Zhang2020RSS}. Simulators must also capture multiphysics behavior, as tactile sensors are cross-modal: for instance, the BioTac transduces skin deformations to fluidic impedances. Furthermore, whereas a small parameter set (e.g., camera intrinsics) can describe variation among visual sensors, no equivalent exists for tactile sensors. Due to manufacturing variability, even sensors of the same type can behave disparately\cite{Narang2020RSS, zapata20203d}.

In this work, we address forward simulation and sim-to-real transfer for the BioTac (Fig.~1). We first develop 3D finite element method (FEM) simulations of the BioTac using a GPU-based robotics simulator\cite{IsaacGym}; the FEM simulations predict contact forces and deformation fields for the sensor for arbitrary contact interactions.
These simulations are designed to reproduce our previous results\cite{Narang2020RSS, Narang2021arXiv}, which utilized an industry-standard, commercial, CPU-based simulator and were carefully validated against real-world experiments. However, the new simulator is freely accessible, and the simulations execute 75x faster.

We then map FEM output to real-world BioTac electrical signals by leveraging recent methods in self-supervised representation learning. Specifically, we collect a large unlabeled dataset of sensor deformation fields from simulation, as well as a smaller dataset of electrical signals from real-world experiments; we then learn latent representations for each modality using variational autoencoders (VAE) \cite{Kingma2014, COMA:ECCV18}. Next, we learn a cross-modal projection between the latent spaces using a small amount of supervised data. We demonstrate that this learned latent projection allows us to accurately predict BioTac electrical signals from simulated deformation fields for unseen contact interactions, including unseen objects. We can also execute the inverse mapping (from signals to deformations) with higher fidelity than in \cite{Narang2020RSS, Narang2021arXiv}, illustrated via a contact-patch estimator.

To summarize, our key contributions are the following:

\begin{enumerate}
    \item An accurate, efficient, and freely-accessible 3D FEM-based simulation model for the BioTac
    \item A novel application of self-supervision and learned latent-space projections for facilitating cross-modal transfer between FEM and electrical data
    \item Demonstrations of sim-to-real transfer through accurate synthesis of BioTac electrical data and estimation of contact patches, both for unseen contact interactions
\end{enumerate}

The simulation model, as well as additional implementation details and analyses, will be posted on our website.\footnote{\url{https://sites.google.com/nvidia.com/tactiledata2}}

\section{RELATED WORKS}

In this section, we review research efforts in sim-to-real transfer, self-supervision, and cross-modal transfer involving tactile sensors that are widely used in current robotics projects. For a recent comprehensive review, see~\cite{li2020review}.

\subsection{Sim-to-Real for Vision-Based Tactile Sensors}

In \cite{Gomes2019ICRA}, visual output of the GelSight\cite{Yuan2017Sensors} was simulated using a depth camera in Gazebo\cite{Koenig2004IROS}, a calibrated reflectance model, and blurring to approximate gel contact. Quantitative evaluation was limited. In \cite{Sferrazza2020IROS}, a custom marker-based tactile sensor\cite{Sferrazza2019Sensors} was simulated using FEM, optics models, and synthetic noise. A U-net was trained on the synthetic data to regress to contact force fields with a resultant error of $0.14 N$; almost all training and testing was conducted on 1 normally-oriented indenter. In \cite{Ding2020ICRA}, the pin locations of the TacTip\cite{Ward2018SoRo} were simulated using an approximate deformation model in Unity\cite{Juliani2020ArXiv}. Parameters were tuned for a plausible baseline, and domain randomization was performed. A fully connected network (FCN) was trained on synthetic data to regress to contact location and angle with a minimum error of $0.5$-$0.7 mm$ and $0.25 rad$. The sensor examined was large, and contact was made over limited orientations.

In comparison to these studies, we use a compact sensor, accurately simulate contact and deformation, do not perform domain adaptation beyond projection, and conduct simulations and experiments on 17 objects over diverse kinematics.

\subsection{Sim-to-Real for Non-Vision-Based Tactile Sensors}

In \cite{Wu2019CoRL}, the electrical output of the  BarrettHand\cite{BarrettHand} capacitive sensor array was simulated using soft contact in PyBullet\cite{Coumans2019PyBullet}. RL policies for stable grasps were trained on the simulator and transferred to the real world. Binary thresholding was applied to tactile signals, limiting precision. In \cite{Lee2020ICRA}, an electrical resistance tomography sensor was simulated using a simplified FEM deformation model, an empirical piezoresistive model, and an FEM conductivity model. FCNs were trained on synthetic data to predict discrete contact location and resultant force, with an $82\%$ success rate for contact and a mean force error of $0.51 N$. For unseen contact scenarios, the error increased to $5.0 N$.

Finally, for the BioTac sensor, multiple studies have addressed the inverse problem of converting electrical output to physical quantities, such as contact location\cite{Lin2013TechRep, Narang2020RSS, Narang2021arXiv}, force\cite{Lin2013TechRep, Su2015Humanoids, Sundaralingam2019ICRA, Narang2020RSS, Narang2021arXiv}, and deformation\cite{Narang2020RSS, Narang2021arXiv}. In particular, our previous work\cite{Narang2020RSS} presented a 3D FEM model of the BioTac, which was built with the industry-standard, CPU-based simulator ANSYS\cite{ANSYS} and carefully validated against experimental data. Electrode signals were then mapped to simulator outputs via PointNet++\cite{Qi2017ArXiv}. The simulations were slow ($7 min$ each on 6 CPUs). In addition, the forward problem of synthesizing electrical output was not addressed; some progress was made in our extension\cite{Narang2021arXiv}.

The forward problem has been further explored in \cite{Ruppel2019IAS, zapata20203d, zapata2020prediction}. In \cite{Ruppel2019IAS}, the BioTac was simulated with an approximate contact model in Gazebo. A deep network was trained to regress from estimated force and real-world contact location to electrical outputs. An existing location estimator was tested on synthetic and real-world data with a mean difference of $0.83 mm$. Training and testing was conducted only on 1 spherical indenter. In \cite{zapata20203d}, PointNet was used to regress from RGB-D images and grasp parameters to tactile readings, and in \cite{zapata2020prediction}, semi-supervised learning was applied. Electrode prediction errors were relatively high for both the simple case of unseen \textit{samples}, as well as the difficult case of unseen \textit{objects}; numerical comparison is provided later.

In contrast to the preceding works, we develop an efficient FEM model, conduct simulations and experiments with 17 objects, regress to continuous electrical signals, and demonstrate accurate predictions for unseen objects.

\subsection{Tactile Self-Supervision and Cross-Modal Transfer}

Numerous studies in robotics have established the utility of multimodal data in task-specific learning. For example, in \cite{Calandra2017CoRL, Calandra2018RAL}, vision and tactile sensing were combined to predict grasp outcomes and select adjustments, and in \cite{Fazeli2019SciRob}, vision and force/torque (F/T) sensing were combined to perform manipulation tasks. In addition, in \cite{lee2019_multimodal}, vision, F/T sensing, and proprioception were used to learn a joint latent space via self-supervision with autoencoders; the output served as perception input to an RL agent for peg-in-hole.

Simultaneously, recent works both outside and within robotics have investigated cross-modal transfer. In \cite{ngiam2011multimodal}, audio-video transfer was performed by learning a shared representation via a bimodal autoencoder; networks trained on one modality were then able to classify the other. In \cite{Chen2017Thematic}, audio-image transfer was achieved via generative adversial networks (GAN). In \cite{LeeLuo2019ICRA, Li2019CVPR}, cross-modal transfer was performed between data from cameras and vision-based tactile sensors, and in \cite{zapata20203d, zapata2020prediction}, transfer was achieved between cameras and output from the BioTac SP.

Finally, previous efforts have applied distinct network architectures to encode BioTac-specific data. In our previous work \cite{Sundaralingam2019ICRA, Narang2020RSS, Narang2021arXiv}, a 3D voxel-grid network, PointNet++, and an FCN were implemented to encode BioTac electrode data and regress to physical quantities such as forces and deformations. In \cite{gutierrez_tactilecnn}, a 2D CNN was used to predict tactile directionality, and in \cite{garcia2019tactilegcn}, a graph convolutional network (GCN) was used to predict grasp stability.

We draw upon the preceding works with some distinctions. Analogous to \cite{ngiam2011multimodal}, we learn latent representations of FEM and BioTac electrical data via self-supervision with autoencoders for cross-modal transfer. Like \cite{lee2019_multimodal}, we learn modality-specific representations, reducing training time and eliminating zero-inputs for non-present modalities. Unlike both, we never formulate a joint representation, but instead learn a projection between the latent spaces using a small amount of supervised data. To encode BioTac electrical data, we use VAEs, as for vision-based tactile sensors in \cite{Lambeta2020ICRA}.

\section{METHODS}

In this section, we first discuss our 3D FEM model, which predicts BioTac contact forces and deformation fields for arbitrary contact interactions. We then discuss our implementation of self-supervision and latent-space projection, which can synthesize BioTac electrical output from unlabeled FEM output and predict contact patches from electrical input. Finally, we describe the simulations and experiments used to collect the FEM and electrical data used in this paper.

\subsection{Finite Element Modeling}
\label{sec:gym_fe_model}

FEM is a variational numerical formulation that 1) divides complex global geometries into simple subdomains, 2) solves the weak form of the governing PDEs in each subdomain, and 3) assembles the solutions into a global one. In 3D FEM, objects are represented as volumetric meshes, which consist of 3D elements (e.g., tetrahedrons or hexahedrons) and their associated nodes. With careful model design, high-quality meshes, and small timesteps, FEM predictions for deformable bodies can be exceptionally accurate \cite{Reddy2019Book, Narang2018AFM}.

In this work, 3D FEM was performed using NVIDIA's GPU-based Isaac Gym simulator\cite{IsaacGym}. Isaac Gym models the internal dynamics of deformable bodies using a co-rotational linear-elastic constitutive model; these bodies interact with external rigid objects via an isotropic Coulomb contact model \cite{Stewart2000SIAM}. The resulting nonlinear complementarity problem is solved via implicit time integration using a GPU-based Newton solver \cite{Macklin2019TOG}. At each timestep, Isaac Gym returns nodal positions (i.e., deformation fields), nodal contact forces (used to compute resultant forces), and element stress tensors.

\begin{figure}
  \centering
  \includegraphics[width=0.95\columnwidth]{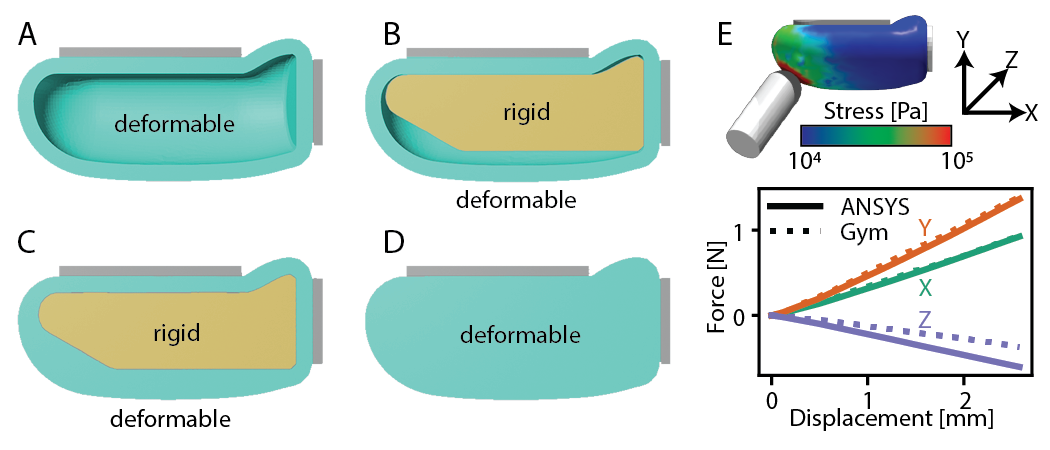}
  \caption{A-D) 4 geometric configurations of the BioTac. Cross-sections are shown. A) Deformable shell, modeling the real-world rubber skin. B) Deformable shell with rigid core, modeling the skin and real-world plastic core. C) Deformable solid with rigid core. D) Deformable solid. Grey rigid bodies were used to apply fixed boundary conditions. E) Optimization results for FEM model. The material parameters of representation B were tuned in Isaac Gym to reproduce the force profile of a shear-rich indentation from ANSYS in \cite{Narang2020RSS, Narang2021arXiv}. The von Mises stress distribution is visualized for the indentation midpoint. The mean $\ell^2$-norm of the force error vector was $0.125 N$. Indenter displacement is relative to the point of initial contact.}
  \label{fig:biotac_geom_approx}
  \label{fig:gym_param_optim}
  \vspace{-8pt}
\end{figure}

To create the FEM model for the BioTac, high-resolution triangular meshes for the external and internal surfaces of the BioTac skin were first extracted from the ANSYS model in \cite{Narang2020RSS, Narang2021arXiv} and simplified via quadric edge collapse decimation in MeshLab\cite{Cignoni2008Euro}. A volumetric mesh was then generated with fTetWild\cite{Hu2020TOG}; similar to \cite{Narang2020RSS, Narang2021arXiv}, the mesh had $\approx 4000$ nodes. Fixed boundary conditions (BCs) were applied to 2 sides of the skin to model the BioTac nail and clamp, respectively; these BCs were implemented by introducing thin rigid bodies at the corresponding locations (visible in Fig. \ref{fig:biotac_geom_approx}A-D), which were attached to adjacent nodes on the skin. External rigid objects (e.g., indenters) were driven into the BioTac via a stiff proportional controller.

Relative to the experimentally-validated ANSYS model in \cite{Narang2020RSS, Narang2021arXiv}, the Isaac Gym model makes 3 critical approximations: 1) collisions are resolved via boundary-layer expanded meshes\cite{Hauser2020Springer}, rather than a normal-Lagrange method, 2) a compressible linear-elastic model is used for the skin, rather than a Neo-Hookean model\cite{Ogden2013Book}, 3) the internal fluid is not modeled. The effects of the first approximation are mitigated by using small collision thicknesses and timesteps ($1\mathrm{e}{-4}s$). However, the second and third approximations are mitigated by endowing the Isaac Gym model with sufficient expressivity and optimizing it to reproduce ANSYS results.

Specifically, 4 distinct geometric configurations of the BioTac were considered in Isaac Gym (Fig. \ref{fig:biotac_geom_approx}A-D). The material properties (elastic modulus $E$, Poisson's ratio $\nu$, and friction $\mu$ of the BioTac skin) were designated as free parameters. In \cite{Narang2020RSS, Narang2021arXiv}, 1 shear-rich indentation of the BioTac was used to calibrate the ANSYS model against real-world data; in this work, the same indentation was resimulated in Isaac Gym and used to calibrate the Gym model against ANSYS data (Fig. \ref{fig:biotac_geom_approx}E). For each configuration, the material properties were optimized via sequential least-squares programming to reproduce the force-deflection profile from ANSYS for the indentation. The cost was defined as the RMS $\ell^2$-norm of the force error vector over time. Subsequently, each optimized configuration was evaluated by resimulating 358 additional indentations from 8 indenters in \cite{Narang2020RSS, Narang2021arXiv}, and comparing the results to the force-deflection profiles from ANSYS.

Among the 4 configurations, the \textit{deformable solid} (Fig. \ref{fig:biotac_geom_approx}D) produced the lowest cost during optimization; however, the \textit{deformable shell with rigid core} (Fig. \ref{fig:biotac_geom_approx}B) produced the lowest cost during evaluation and was thus selected. For this representation, the optimal values of $E$, $\nu$, and $\mu$ were $1.55e6 Pa$, $0.316$, and $0.783$, respectively; the mean $\ell^2$-norm of the force error vector was $0.125N$ (Fig. \ref{fig:gym_param_optim}E). In comparison, the optimal values for the ANSYS model in \cite{Narang2020RSS} were $2.80e5 Pa$, $0.5$, and $0.186$, indicating that the Isaac Gym model compensated for its linearity and lack of fluid by increasing elastic modulus, compressibility, and friction.

\subsection{Learning Latent Space Projections}
\label{sec:repr_learning}

\begin{figure}
  \centering
  \includegraphics[width=0.9\columnwidth]{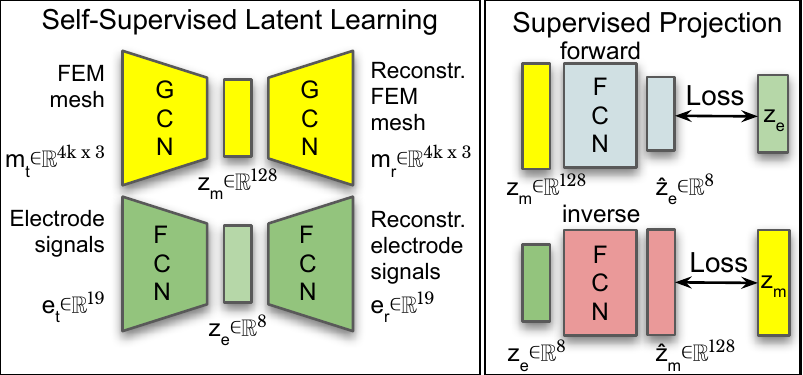}
  \caption{Learning structure. To map between FEM deformations and BioTac electrode signals, modality-specific latent representations were learned via self-supervision. Specifically, graph convolutional networks (GCN) compressed deformed meshes with 4000 nodes to a 128-dim. latent space, and fully-connected networks~(FCN) compressed the 19 electrode signals to an 8-dim. latent space. Next, FCNs were used on a small supervised dataset to learn forward and inverse projections between the latent spaces.}
  \label{fig:learning_arch}
  \vspace{-8pt}
\end{figure}

Although FEM captures the effects of contact on BioTac deformations, the BioTac then transduces deformations to fluidic impedances measured at electrodes. Simulating the mapping between deformations and impedances is complex; thus, this mapping was learned. Specifically, modality-specific latent representations were learned using self-supervision, facilitating compression, mitigating noise, and reducing overfitting. Projections were then learned between the latent spaces, enabling cross-modal transfer (Fig.~\ref{fig:learning_arch}).

To learn a latent representation for the FEM deformations, 
convolutional mesh autoencoders from \cite{COMA:ECCV18} were trained, which applied graph convolutional networks (GCN) and reduced the mesh data from 4000 nodal positions~($\mathbb{R}^{4k\times 3}$) to 128 units. To learn a latent representation for real-world BioTac electrode signals, an autoencoder composed of FCNs was trained, which reduced the electrode data from 19 impedances to 8 units. Latent dimensions were chosen via hyperparameter search. Both networks were trained as VAEs to generate smooth mappings to the latent space\cite{Kingma2014}.

To learn the projections between the latent spaces, 2 FCNs were trained. The first network projected forward from the FEM mesh latent space~$z_m$ to the BioTac electrode latent space~$z_e$, whereas the second network projected inversely from~$z_e$ to~$z_m$. During training, the previously-described VAEs were frozen and provided with supervised data from \cite{Narang2020RSS}, generating latent pairs of $z_m$ and $z_e$. These pairs were used to train the projection networks with an RMS loss. Without freezing the VAEs, the networks overfit.

In the FEM deformation VAE, the encoder consisted of an initial convolution with filter size 128, 4~``convolve+downsample'' layers with sizes $[128,128,256,64]$ and downsampling factors~$[4,4,4,2]$, a convolution with size 64, and 2~fully-connected layers with dimensions $[512,128]$. The decoder was symmetric with the encoder, using ``upsample+convolve'' instead of ``convolve+downsample''. In the BioTac electrode VAE, the encoder consisted of 4~fully connected layers with $[256,128,64,8]$ neurons, respectively. The decoder was again symmetric with the encoder. The forward and inverse projection networks consisted of 3~fully connected layers with $[256,128,128]$ and~$[128,128,256]$ neurons, respectively, and dropout of $0.3$. Exponential linear unit (ELU) activations were applied, and the Adam optimizer was used with initial learning rate of $1e$-$3$ and decay of $0.95$.

\subsection{Dataset Collection}
\label{sec:data_collect}

\begin{figure}
  \centering
  \includegraphics[width=0.98\columnwidth]{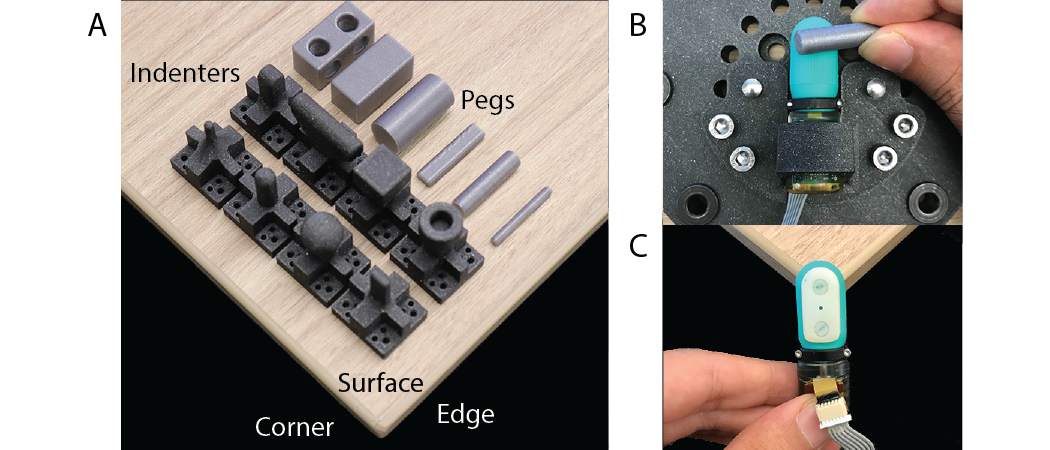}
  \caption{Data collection. A) 17 objects and table features were used to generate data in simulation and the real world. The objects were designed to mimic diverse physical features (e.g., corners, edges, buttons, holes). B-C) The BioTac was constrained when applying kinematically-randomized indentations with the pegs, but unconstrained when contacting table features.}
  \label{fig:exp_data_collect}
  \vspace{-8pt}
\end{figure}

For the preceding learning steps, data was collected in both simulation and the real world. For learning the latent representations, unlabeled mesh data was collected by simulating kinematically-randomized interactions on the optimized BioTac model with 6 pegs and 3 table features (surfaces, edges, and corners) in Isaac Gym. Unlabeled experimental electrode data was collected by manually indenting these objects in the real world. For learning the latent projections, labeled data was collected by exactly resimulating 359 indentations from \cite{Narang2020RSS, Narang2021arXiv} on the optimized BioTac model in Isaac Gym (as stated in Sec.~\ref{sec:gym_fe_model}); these were directly matched to corresponding experimental electrode data in the dataset from \cite{Narang2020RSS, Narang2021arXiv}. 72\% of contact interactions were allocated for training, 18\% for validation, and 10\% for testing. 

In total, 2.6k unique contact interactions were executed and 50k timesteps of FEM data were sampled. All objects used in simulation and experiments are shown in Fig.~\ref{fig:exp_data_collect}. Data collection visualizations are in the supplementary video.

\section{Experiments \& Results}
\label{sec:results}

\begin{figure*}
  \centering
\includegraphics[width=0.95\textwidth]{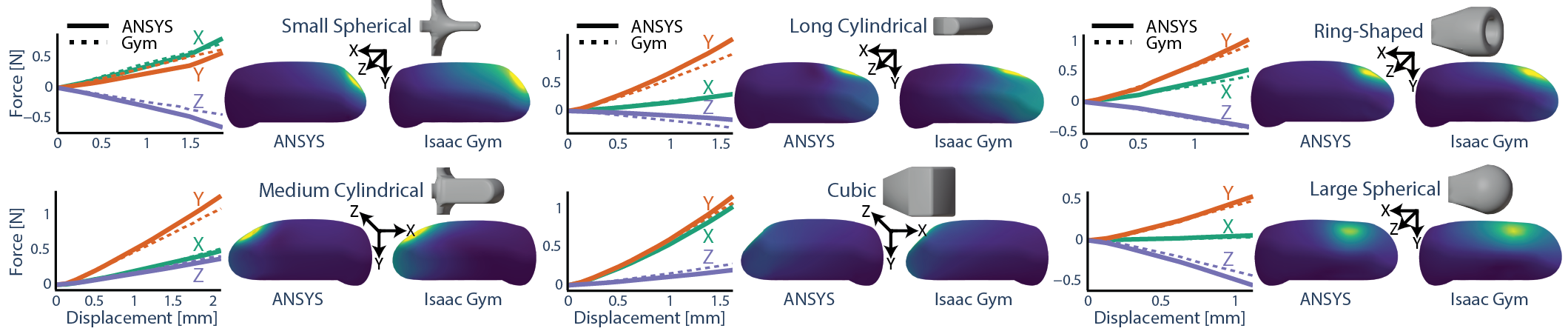}
  \caption{Validation of FEM model. The optimized model from Isaac Gym was compared to an experimentally-validated model from ANSYS. Force-deflection profiles, as well as nodal deformation fields at maximum indentation depth, are illustrated for 6 randomly-selected indentations with 6 different indenters. Nodal deformation fields are colored according to corresponding displacements (i.e., change in nodal positions relative to the no-contact state). The 2 not-pictured indenters are a medium-sized spherical indenter and a small cylindrical indenter. Larger, higher-contrast graphics are on our website.}
  \label{fig:gym_ansys_comp}
\end{figure*}

In this section, we present our results on FEM validation, synthesis of BioTac electrode signals from FEM deformations, and contact patch estimation from electrode signals.

\begin{figure*}[b]
  \centering
  \includegraphics[trim=0 0.cm 0 0cm,clip,width=0.996\textwidth]{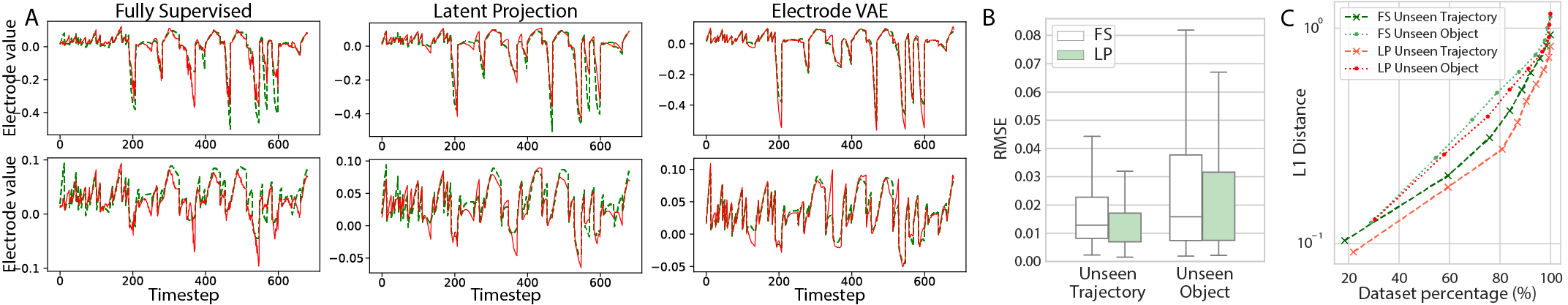}
    \caption{Electrode prediction results. A) Visual comparisons for 2 high-magnitude, high-variation electrodes during contact interactions with an unseen object. Electrode range corresponds to a force range of $0.5$-$19.8N$. Each peak corresponds to a distinct interaction. Green and red lines indicate ground-truth and prediction, respectively. Our latent projection (LP) approach can predict peaks more accurately than a fully-supervised (FS) baseline, and often outperforms the VAE used in training. B) RMS error over all electrodes and interactions, for unseen trajectories and objects. Our approach has lower median errors and interquartile ranges. C) Coverage plot, with $\ell^1$ distance to ground-truth. Our approach has lower errors over nearly the full data distribution.}
  \label{fig:elec_recon}
  \vspace{0pt}
\end{figure*}

\subsection{FEM Validation}

As described earlier, the force-deflection profiles produced by the optimized model in Isaac Gym were compared to those produced by ANSYS over 358 indentations distributed across 8 indenters (Fig. \ref{fig:gym_ansys_comp}). The mean $\ell^2$-norm of the force error vectors over all indentations ranged from $0.0876 N$ for a medium-sized cylindrical indenter (\textit{less} than the training error of $0.125 N$) to $0.259 N$ for a medium-sized spherical indenter, with a mean of $0.153 N$ across all indenters. Thus, despite being optimized using force-deflection data from only a single indentation, the Isaac Gym model strongly generalized across a diverse range of objects and indentations.

The corresponding FEM deformation fields (i.e., the nodal positions of the deformed FEM meshes) were also compared between Isaac Gym and ANSYS (Fig. \ref{fig:gym_ansys_comp}). For each dataset, the maximum and mean $\ell^2$-norms of the nodal displacement vectors were computed over all indentations. The mean error between the maximum norms across all indenters was $1.41\mathrm{e}{-4} m$, and the mean error between the mean norms was $1.81\mathrm{e}{-5} m$. Thus, again, the Isaac Gym model was shown to strongly generalize. These low errors were particularly important, as the nodal deformation fields from Isaac Gym were used as input for subsequent learning.

Finally, simulation speed was compared between the Isaac Gym and ANSYS models. The total simulation time for all 359 indentations was approximately 42 hours  (7.08 minutes per sim) on 6 CPUs in ANSYS, but 33 minutes (5.57 seconds per sim) using 8 parallel environments (1 per indenter) on 1 GPU in Isaac Gym. For clarity, Isaac Gym can only currently simulate deformable solids with a linear-elastic model and linear tetrahedral elements; such a model comprises only a small fraction of those that can be simulated within state-of-the-art FEM software such as ANSYS. However, for the current application, Isaac Gym is highly favorable. 

\subsection{Learning-Based Regression and Estimation}

For regression from FEM deformations to BioTac electrode signals, 2 learning methods were evaluated: 1) the method proposed in this paper, denoted \emph{Latent Projection}, which used unlabeled data for latent representation and labeled data for projection, and 2) a PointNet++ baseline \cite{Qi2017ArXiv}, denoted \emph{Fully Supervised}, which used only labeled data, with 128 nodes sampled from the FEM mesh as in \cite{Narang2020RSS, Narang2021arXiv}. For reference, output is also shown for the \emph{Electrode VAE}, which is used when training \emph{Latent Projection}.

When evaluating generalization to novel objects, networks were trained on all objects \textit{except} the ring (see Fig.~\ref{fig:gym_ansys_comp}) and tested on this indenter; the ring has the most unique (thus, challenging) geometry in the dataset. These experiments are denoted ``Unseen Object.'' (Results for other unseen objects are on our website.) When evaluating generalization to novel contact interactions with seen objects, the trained networks are tested on unseen interactions with the other indenters. These experiments are denoted ``Unseen Trajectory.''

A qualitative comparison of regression results between the learning methods is depicted in Fig.~\ref{fig:elec_recon}A for 2 high-signal, high-variation electrodes over numerous interactions. Raw electrode values were between [0, 4095] digital output units and were tared and normalized to [-1, 1]. For the challenging ``Unseen Object'' case, \emph{Latent Projection} could predict several signal peaks over multiple electrodes that \emph{Fully Supervised} could not capture. Additionally, \emph{Latent Projection} predictions were consistently noise-free, whereas \emph{Fully Supervised} ones exhibited low-magnitude, high-frequency noise. Finally, \emph{Latent Projection} often outperformed \emph{Electrode VAE}, showing the importance of mesh information in electrode signal synthesis. Predictions for the easier ``Unseen Trajectory'' case have higher fidelity and are thus not shown.

Quantitative comparisons between the learning methods are illustrated in Fig.~\ref{fig:elec_recon}B-C. RMS errors over all electrodes and interactions are compared in Fig.~\ref{fig:elec_recon}B. The \emph{Latent Projection} approach performs better than \emph{Fully Supervised} for both ``Unseen Trajectory'' and ``Unseen Object,'' in terms of both median error and interquartile range. Median errors for ``Unseen Trajectory'' were \emph{Fully Supervised}:~$0.012$ (25 raw units) and \emph{Latent Projection}:~$0.010$ (20 units), and those for ``Unseen Object'' were $0.016$ (32 units) and $0.015$ (31 units), respectively. Our errors are substantially lower than the errors from~\cite{zapata2020prediction}, which were~$195$ units for an easier case of unseen \textit{samples}, and~$305$ units for unseen objects. A coverage plot is shown in Fig.~\ref{fig:elec_recon}C, where the~$\ell^1$ distance to the ground-truth electrode signals is depicted. For both ``Unseen Trajectory'' and ``Unseen Object,'' \emph{Latent Projection} outperforms \emph{Fully Supervised} for nearly the entire distribution of data.

\begin{figure}
  \centering
  \includegraphics[trim= 0 1.6cm 0cm 0cm ,clip,width=0.85\columnwidth]{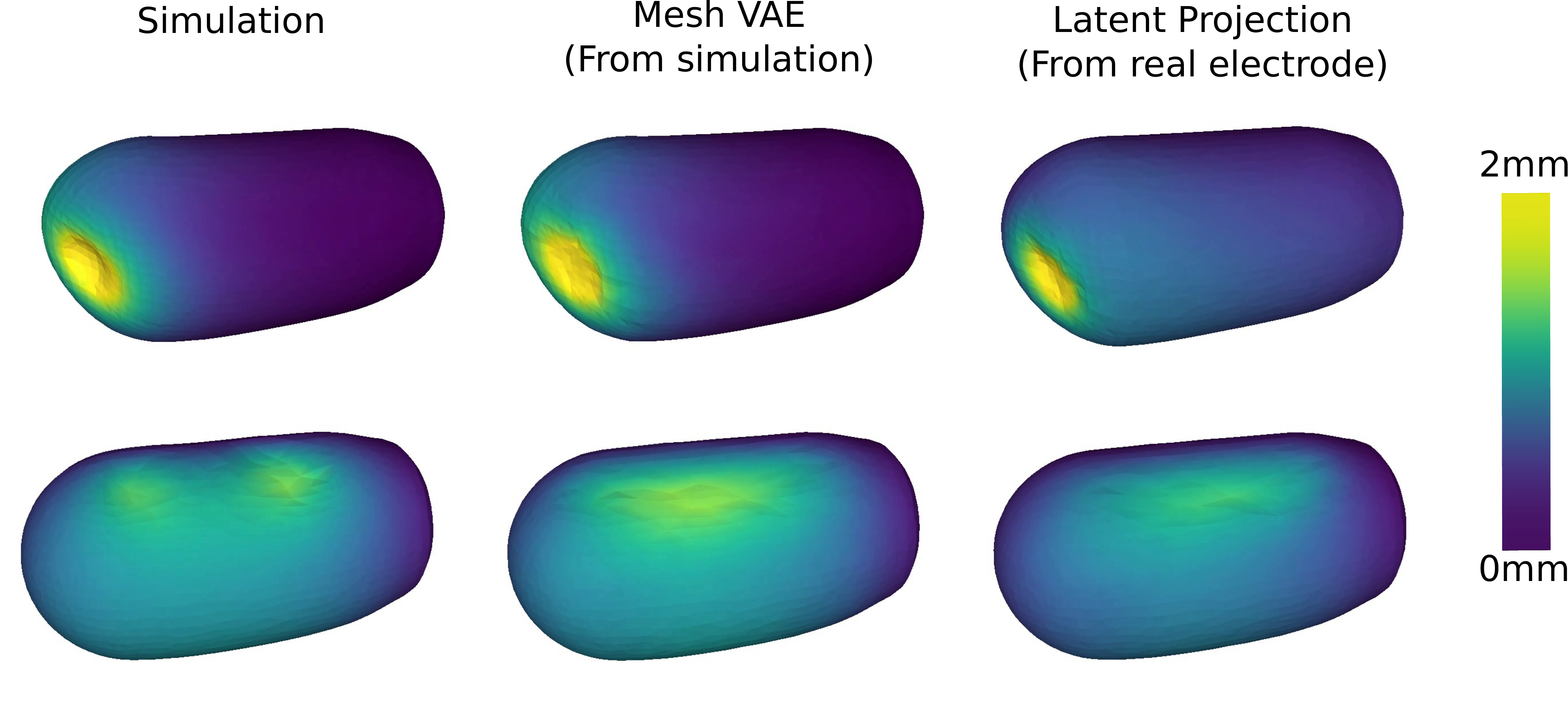}
  \caption{Predicted FEM deformations for unseen trajectories and objects. The columns show the raw FEM output, the output of the mesh VAE, and the prediction from real-world electrode signals through our latent projection. The colorbar shows the Euclidean distance from the undeformed state. Top) ``Unseen Trajectory'' case, for a randomly-selected indentation. Predictions are consistently accurate. Bottom) ``Unseen Object'' case, for a randomly-selected indentation with the most distinct indenter, the ring. Predictions do not capture the bimodal deformation distribution due to limitations of the VAE, which has not seen any examples of such distributions in training.}
  \label{fig:mesh_recon}
  \vspace{-8pt}
\end{figure}

For regression from BioTac electrode signals to FEM deformations, we again evaluate \emph{Latent Projection}. For reference, we also show output of the FEM \emph{Mesh VAE}, which is used when training \emph{Latent Projection}. A visual comparison is shown against ground-truth FEM output in Fig.~\ref{fig:mesh_recon} for random indentations. For ``Unseen Trajectory'', \emph{Latent Projection} consistently predicted ground-truth, capturing deformation magnitudes and distributions. For ``Unseen Object'', \emph{Latent Projection} consistently captured magnitudes and distributions, but not bimodality. As seen from \emph{Mesh VAE}, this limit is due to the mesh autoencoder (specifically, bottleneck size) rather than the projection. As before, we only show results for the most challenging unseen object, the ring; strong performance on other objects is shown in the video.

\begin{figure}
  \centering
  \includegraphics[trim=0 0.3cm 0 0,clip,width=0.97\columnwidth]{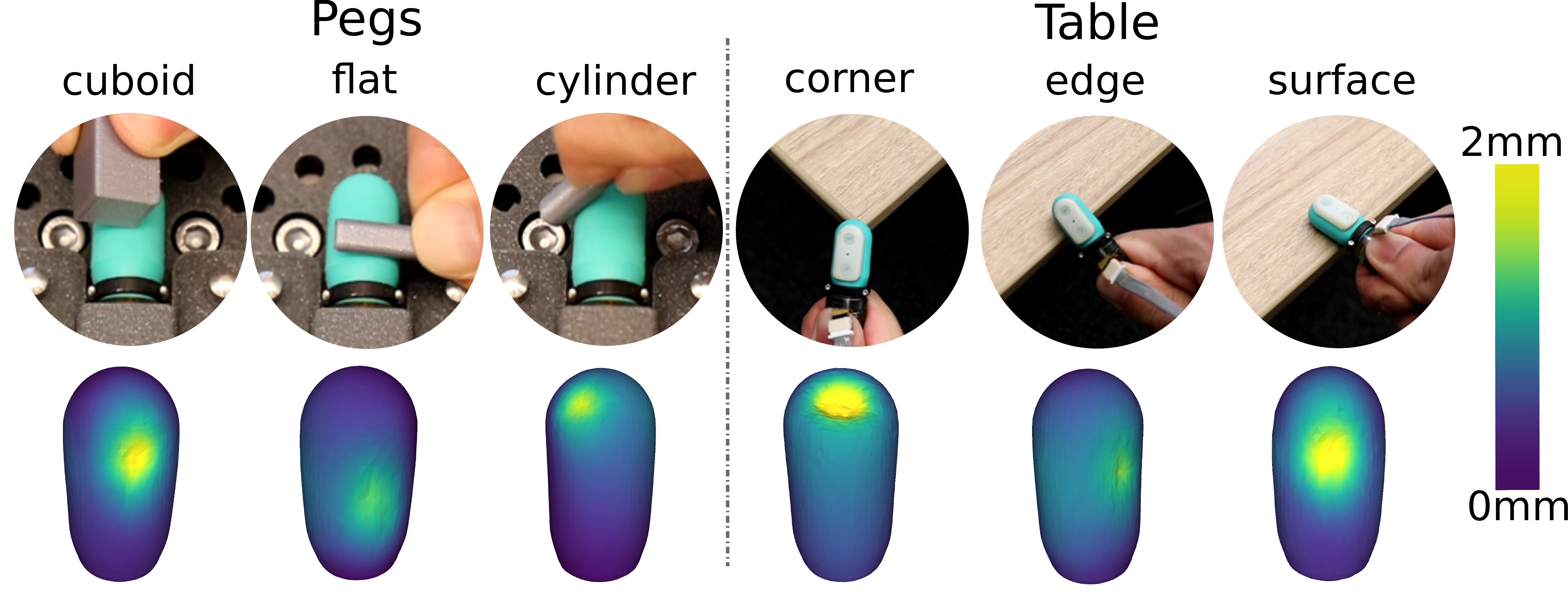}
  \caption{Contact patch estimation for free-form interactions. The colorbar shows the Euclidean distance from the undeformed state. Top) Real-world contact with various unseen objects and table features. Bottom) Predicted contact patches using our learned latent projections. Contact patches match contacting features; for example, interactions with the corners of the flat peg and table produce highly-localized, high-magnitude deformation.}
  \label{fig:peg_contact}
  \vspace{-8pt}
\end{figure}

As a final demonstration, we also conducted free-form interactions of the BioTac with unseen objects and visualized the estimated contact patches (Fig.~\ref{fig:peg_contact}). For diverse pegs and table features, predicted deformations were visually accurate. For instance, contact patch locations were accurately predicted across the full spatial limits of the BioTac, and interactions with the corners of a cuboid peg and table showed high-magnitude, highly-localized patch deformations.

\section{Discussion}
In this paper, we present a framework for synthesizing BioTac electrical signals and estimating contact patches for novel contact interactions. The framework consists of 1) a 3D FEM model, which simulates contact between the BioTac and objects and outputs BioTac deformation fields, 2) VAEs that output compressed representations of the deformation fields and electrode signals, and 3) projection networks that perform cross-modal transfer between the representations to facilitate regression of electrode signals or contact patches.

This work has several key contributions. First, compared to our previously-presented, experimentally-validated FEM model \cite{Narang2020RSS}, the current model is nearly equivalent, available in an open-access robotics simulator, and 75x faster. Second, our work presents one of the first applications of cross-modal self-supervision for tactile sensing; we show that this approach outperforms supervised-only methods for regressing to BioTac electrical signals. Third, for the first time, we accurately predict these signals for unseen interactions, including unseen objects. Finally, we can reconstruct BioTac deformations from real electrical signals with high fidelity. 

The present study also has limitations. First, although our FEM model is substantially faster than previous efforts, it currently takes approximately $5.6 s$ to simulate $6 mm$ of indentation, which prohibits dynamic model-predictive control applications. Furthermore, although our networks accurately predicted electrode signals for unseen trajectories and objects, evaluation was performed for 1 BioTac; to compensate for manufacturing variation, unlabeled and labeled data from more BioTacs may be necessary to fine-tune the VAEs and projection networks. Future work will focus on applying our simulation and learning framework to non-BioTac sensors. In the process, we aim to develop powerful, generalized representations of tactile data that can serve as the foundation for transfer learning across sensors of entirely different modalities, such as the BioTac and GelSight.


\section*{ACKNOWLEDGMENT}

We thank V. Makoviychuk, K. Guo, and A. Bakshi for their collaboration with Isaac Gym, as well as K. Van Wyk, A. Handa, and T. Hermans for their feedback.


\bibliographystyle{IEEEtran}
\bibliography{refs.bib}

\end{document}